\documentclass[conference]{IEEEtran}
\IEEEoverridecommandlockouts

\usepackage{cite}
\usepackage{amsmath,amssymb,amsfonts}
\usepackage{algorithmic}
\usepackage{graphicx}
\usepackage{textcomp}
\usepackage{xcolor}
\usepackage{hyperref}
\usepackage{booktabs}

\def\BibTeX{{\rm B\kern-.05em{\sc i\kern-.025em b}\kern-.08em
    T\kern-.1667em\lower.7ex\hbox{E}\kern-.125emX}}
\begin{document}

\title{A Dual-Stream Physics-Augmented Unsupervised Architecture for Runtime Embedded Vehicle Health Monitoring\\
\thanks{This research was conducted within the MOVER Program, which aims to drive decarbonization, technological innovation, and competitiveness in the Brazilian automotive sector through research projects focused on efficient machine learning and embedded systems. Therefore this work was partially supported by FUNDEP grants Rota 2030/Linha VI 29271.02.01/2022.01-00 and 29271.03.01/2023.04-00.}
}

\author{\IEEEauthorblockN{1\textsuperscript{st} Enzo Nicolás Spotorno, 2\textsuperscript{nd} Antônio Augusto Fröhlich}
\IEEEauthorblockA{\textit{Software/Hardware Integration Lab - Department of Informatics and Statistics} \\
\textit{Federal University of Santa Catarina (UFSC)}\\
Florianópolis, Brazil \\
\{enzoniko, guto\}@lisha.ufsc.br}
}

\maketitle

\begin{abstract}
Runtime quantification of vehicle operational intensity is essential for predictive maintenance and condition monitoring in commercial and heavy-duty fleets. Traditional metrics like mileage fail to capture mechanical burden, while unsupervised deep learning models detect statistical anomalies, typically transient surface shocks, but often conflate statistical stability with mechanical rest. We identify this as a critical blind spot: high-load steady states, such as hill climbing with heavy payloads, appear statistically normal yet impose significant drivetrain fatigue. To resolve this, we propose a Dual-Stream Architecture that fuses unsupervised learning for surface anomaly detection with macroscopic physics proxies for cumulative load estimation. This approach leverages low-frequency sensor data to generate a multi-dimensional health vector, distinguishing between dynamic hazards and sustained mechanical effort. Validated on a RISC-V embedded platform, the architecture demonstrates low computational overhead, enabling comprehensive, edge-based health monitoring on resource-constrained ECUs without the latency or bandwidth costs of cloud-based monitoring.
\end{abstract}


\section{Introduction}

The effective management of modern vehicle fleets (spanning commercial logistics, municipal services, and heavy-duty operations) depends on a precise understanding of vehicle utilization and mechanical wear. For decades, operators have relied on simplistic metrics such as mileage or hours of operation to estimate intensity and schedule maintenance \cite{kazemian2024sensor}. However, these coarse measures fail to reflect the true physical burden on vehicle components \cite{Wong2022}, resulting in suboptimal maintenance, inequitable cost allocation, and limited insights into operational efficiency.

Advances in edge-based Information and Communication Technology (ICT) now allow for granular, runtime analysis of sensor streams, deploying the decentralized AI necessary for sustainable, smart transport without the bandwidth costs of continuous cloud transmission. In parallel, unsupervised deep learning models, such as LSTM Autoencoders, have become prevalent for anomaly detection in time-series telemetry \cite{zamanzadeh2024deep}. These models learn patterns of ``normal'' driving and flag deviations (e.g., pothole impacts, surface shocks, sharp cornering) as potential risks. In previous work \cite{spotorno2025billing}, we analyzed the operational trade-offs between prediction accuracy, data-label efficiency, and
computational cost for deploying these models on embedded hardware, by profiling these methods on a RISC-V platform. We conducted a comparative analysis of distinct architectures across a spectrum of supervision: unsupervised reconstruction-based heuristics, semi-supervised clustering-based methods, and a fully supervised XGBoost benchmark. The study revealed that while the supervised benchmark achieved the highest predictive fidelity, it incurred a significant computational penalty, consuming approximately $22.1\,\mu\text{J}$ per inference compared to just $0.67\,\mu\text{J}$ for the heuristic baseline. This 30-fold increase in energy cost, coupled with execution times reaching $200\,\mu\text{s}$ for clustering-based methods, confirmed that unsupervised architectures offer a necessary compromise for resource-constrained ECUs.

However, this computational characterization highlighted a functional blind spot: \textbf{the decoupling of statistical anomaly scores from physical load}. Black-box models operate purely on signal deviation, implicitly assuming that statistical normality equates to mechanical rest. However, high-stress but predictable operations, such as a heavy truck maintaining constant speed up a steep grade, result in smooth, consistent signals. These operations are statistically ``normal'', yielding low reconstruction error \cite{walczyna2024enhancing}, yet impose significant drivetrain fatigue. This disconnect hinders accurate wear modeling and precise fatigue estimation. Measuring actual mechanical load provides a more reliable indication of component fatigue than static mileage logs. On embedded ECUs, achieving this demands lightweight, label-efficient methods that can operate without the addition of expensive physical sensors (e.g., torque transducers).

Pure data-driven anomaly detection captures statistical deviations but not sustained physical effort or underlying mechanical work (e.g., force, energy, or power over time), motivating hybrid approaches that incorporate physical knowledge alongside data analytics to improve reliability and interpretability in predictive maintenance of mechanical systems. Literature on hybrid physics-based and data-driven modeling shows that integrating physical insights with statistical models enhances robustness in regimes with limited data \cite{raissi2019physics, han2025physics}, which is analogous to maintenance contexts where purely data-driven methods struggle with generalization and physical consistency. Microscopic physics models demand unavailable high-fidelity data. We utilize \textit{macroscopic physics proxies}, which are robust, computable approximations from standard sensors, that complement unsupervised learning without loss-function constraints.

The main contributions of this work are threefold: first, we quantifiably demonstrate the limitations of standard unsupervised models in capturing steady-state mechanical work, evidenced by a near-zero correlation with drivetrain load proxies; second, we characterize the bias of reconstruction-based detectors toward high-frequency lateral dynamics; and third, we propose a lightweight Dual-Stream Architecture that fuses unsupervised anomaly detection with macroscopic physics proxies. This hybrid approach enables multi-dimensional, explainable health monitoring on resource-constrained embedded hardware.

The remainder of this paper is organized as follows: Section~\ref{sec:related} reviews related work. Section~\ref{sec:methods} details the proposed Dual-Stream Architecture, including the unsupervised baseline, macroscopic physics proxies, and fusion logic. Section~\ref{sec:setup} describes the experimental setup and embedded validation platform. Section~\ref{sec:results} presents the experimental evaluation, analyzing the decoupling of anomaly scores from load and demonstrating the efficacy of the fused health vector. Finally, Section ~\ref{sec:discussion_conclusion} discuss implications and future directions.

\section{Related Work}
\label{sec:related}

This section overviews prior research across four domains (cost allocation methods, anomaly detection and state estimation, physics-informed learning, and computational constraints in embedded systems), motivating operational effort indices for vehicles.

A primary application driving the need for operational effort indices is Activity-Based Costing (ABC), which allocates costs according to actual resource consumption rather than fixed rates \cite{mattetti2022canbus}. The increasing availability of on-board sensor data (e.g., CAN bus) has enabled automatic activity identification for dynamic ABC, but most approaches remain limited to discrete classification with manually assigned costs and therefore lack a continuous quantification of varying loads (for example, payload-dependent hill climbing). Related frameworks such as usage-based insurance incorporate driving-behavior metrics for risk assessment \cite{bian2018good, chan2025assessing}, yet these models typically emphasize safety violations (e.g., hard braking) rather than cumulative mechanical wear, reducing their usefulness for maintenance planning. Likewise, Total Cost of Ownership (TCO) analyses acknowledge that vehicle costs extend beyond purchase price to include fuel, maintenance, insurance, and depreciation \cite{sun2024systemic}, but allocating variable costs accurately is still challenging for mixed-use fleets with diverse operational profiles \cite{da2023gis} when there is no reliable, physics-based metric to quantify the mechanical burden of individual missions.

Vehicle dynamic state estimation and anomaly detection provide much of the technical foundation for on-board monitoring. Classical state estimators recover quantities such as velocity, sideslip, and yaw rate \cite{guo2018vehicle}, while data-driven anomaly detection, particularly reconstruction-based autoencoders, has become popular for monitoring continuous sensor streams \cite{zamanzadeh2024deep, spotorno2025billing}. These methods are effective at filtering volitional driving to highlight discrete impacts (e.g., potholes), but they typically treat statistical normality as benign regardless of the underlying physical work performed, and thus lack the physics-based context needed to estimate cumulative wear. Multi-modal approaches that fuse visual and LiDAR data offer more comprehensive monitoring \cite{cserni2025mm}, yet computational constraints frequently preclude their deployment on-board. 

Traditional physics-based models (PBMs) remain attractive for interpretability and their ability to capture degradation dynamics for control \cite{zagorowska2020survey}, but often require microscopic parameters (e.g., torque, tire–road friction coefficients) that are not available in standard commercial monitoring \cite{arias2021combining}. External alternatives such as GIS-based wear indices avoid the need for on-board sensors but sacrifice runtime granularity \cite{da2023gis}. Time-series techniques like Dynamic Time Warping provide robust similarity measures but incur quadratic computational costs that hinder on-board use \cite{xu2017accelerating}; although recent time-series embedding methods show promise for efficient feature extraction in resource-constrained environments \cite{irani2025time}, a persistent bottleneck across prognostics methods is data annotation, since generating ground-truth wear labels requires structured physical testing protocols that are orders of magnitude more expensive than labeling existing operational data \cite{akrim2023self}.

Integrating physical knowledge into learning systems has been a major trend aimed at bridging data-driven and model-based approaches. Physics-informed neural networks and hybrid architectures inject constraints or energy-based proxies into learning objectives to improve generalization \cite{raissi2019physics}, but in vehicular contexts, these approaches often demand high-fidelity data or computational resources that are unsuitable for embedded deployment. Self-supervised learning methods help mitigate data scarcity in prognostics \cite{akrim2023self}, yet they too can be computationally intensive. Deep metric learning offers a route to few-shot classification of vehicle states when labeled examples are scarce \cite{li2023deep}, although its effectiveness depends on extensive pre-training with representative base classes. On the positive side, recent demonstrations of lightweight energy prediction models for wireless transceivers in embedded automotive systems provide a precedent for conducting feasibility analyses that explicitly consider runtime cost \cite{SpotornoWCNC2025}. Simulation and validation tools such as the CARLA simulator and the SmartData Framework also supply scalable, high-fidelity data-generation pipelines and realistic sensor emulation that are useful for developing and testing physics-aware, deployable models \cite{Dosovitskiy:ACRL:2017, Hoffmann:DAES:2024}.

Finally, the practical constraints of deploying advanced monitoring on resource-limited Electronic Control Units (ECUs) impose stringent requirements on algorithm design \cite{spotorno2025billing}. Runtime processing must be selected to minimize energy consumption and thermal overhead so as not to interfere with critical vehicle control tasks; this creates a clear imperative to balance prediction accuracy against execution efficiency and to ensure feasibility on low-power architectures (for example, RISC-V) while handling continuous sensor streams from multiple sources. These strands collectively motivate physics-grounded, resource-efficient operational effort indices.

\section{Proposed Dual-Stream Architecture}
\label{sec:methods}

The architecture integrates two computational streams to resolve the decoupling between statistical anomalies and physical work. We process standard low-frequency (10\,Hz) inertial/GNSS data ($a_{x,y,z}, v$), segmented into non-overlapping 30-sample (3\,s) windows to capture both transient hazards and macroscopic load. 

\textbf{Stream A (Unsupervised ML)} employs a symmetric 3-layer LSTM Autoencoder ($128 \to 64 \to 32$) trained on smooth-road data (Adam, $\eta=10^{-3}$) to learn ``statistical normality''. The RMS reconstruction error, $\mathrm{A}_{\mathrm{ML}}(t)$, flags high-frequency unpredictability (e.g., potholes) but typically filters smooth, predictable loads \cite{zamanzadeh2024deep, spotorno2025billing}. To complement this, \textbf{Stream B (Physics Proxies)} calculates four macroscopic stress indices \cite{iso2631_1997, Wong2022} over aligned windows. 
\textit{Suspension Stress} ($E_{\mathrm{susp}}$) and \textit{Lateral Stress} ($E_{\mathrm{lat}}$) utilize squared jerk integrals to estimate vibration-induced fatigue and chassis torsion, respectively, motivated by cumulative exposure metrics such as the Vibration Dose Value (VDV) defined in \cite{iso2631_1997}, which aggregates transient shocks over time to quantify severity, avoiding expensive traditional stress-strain analysis and cycle counting \cite{khan2025estimation}, while being lightweight proxies:

\begin{equation}
E_{\mathrm{susp}} = \int \left| \dot{a}_z \right|^2 dt, \quad E_{\mathrm{lat}} = \int \left| \dot{a}_y \right|^2 dt.
\end{equation}

 To ensure embedded feasibility, we apply gravity compensation in the body frame by utilizing the estimated pitch angle ($\theta$) to isolate the true vertical dynamic acceleration ($a_z^{\mathrm{gc}} = a_z - g\cos\theta$) and compute jerk on the 10\,Hz stream via symmetric central differencing \footnote{\label{fn:jerk} Jerk was estimated by a symmetric central difference with $\Delta t=0.1\,$s, and the differentiated signal was smoothed with a short zero-phase 5-point moving average to mitigate numerical amplification of IMU noise at 10\,Hz. Nevertheless, 10\,Hz was chosen to match the target ECU acquisition and because these \emph{macroscopic} proxies target sustained, low-frequency loading (dominant energy $<5\,$Hz); high-frequency micro-vibrations ($>5$\,Hz) are therefore outside the intended scope.}. Here $\theta$ denotes vehicle pitch (inclination) estimated in the body frame; with this convention $g\sin\theta$ gives the longitudinal gravity component and $g\cos\theta$ the vertical component used for gravity compensation.

\textit{Drivetrain Stress} ($P_{\mathrm{drive}}$) captures positive tractive work. Road grade $\sin\theta(t)$ is estimated from the IMU pitch angle, enabling the detection of sustained climbing effort even at constant velocity. Mass $m(t)$ is treated as a mission-specific parameter, while $F_{\mathrm{drag}}$ represents nominal rolling resistance, which is calibrated from typical tire-road coefficients and nominal vehicle weight \cite{Wong2022}: 

\begin{equation}
P_{\mathrm{drive}} = \max\{0, (m (a_x + g \sin\theta) + F_{\mathrm{drag}}) \cdot v\}.
\end{equation}

Finally, \textit{Braking Stress} ($E_{\mathrm{brake}}$) approximates dissipative energy during deceleration phases ($a_x(t) < 0$), serving as a practical engineering proxy for brake system thermal loading that includes service braking contributions without requiring brake pressure signals \cite{Wong2022}. Integrated braking dissipation correlates with thermal load and pad/disc wear and is commonly used as a life‑consumption proxy \cite{voloacua2025motor}: 

\begin{equation}
E_{\mathrm{brake}} = \int \max\{0, -m a_x\} \cdot v \, dt.
\end{equation}

These proxies are robust to parameter uncertainty (e.g., variable mass $m$) \cite{guindani2025truckrss} and computationally lightweight, ensuring temporal alignment with the ML stream for unified health vector generation. Therefore, the architecture fuses the outputs of both streams into a multi-dimensional health vector $H_t$ at each window $t$: $H_t = \big[ \, \mathrm{A}_{\mathrm{ML}}(t), \, \mathrm{W}_{\mathrm{Phys}}(t) \, \big]$, where $\mathrm{A}_{\mathrm{ML}}(t)$ is the reconstruction error from Stream A, and $\mathrm{W}_{\mathrm{Phys}}(t)$ is the normalized aggregate of proxies from Stream B. The final decision logic employs a max-pooling strategy: 

\begin{equation}
\mathrm{Score}(t) = \max(\mathrm{Norm}(\mathrm{A}_{\mathrm{ML}}(t)), \mathrm{Norm}(\mathrm{W}_{\mathrm{Phys}}(t))),
\end{equation}

where $\mathrm{Norm}(\cdot)$ denotes min-max scaling relative to calibration missions. This logic ensures a high health score is triggered if \textit{either} instability is compromised OR work limits are exceeded (Table \ref{tab:logic}). This addresses the blind spot by ensuring that statistically normal but high-load events are captured by the physics component even when the ML stream remains quiescent.

\begin{table}[t]
\caption{Dual-Stream Decision Logic Matrix}
\label{tab:logic}
\centering
\begin{tabular}{@{}llll@{}}
\toprule
Scenario & Stream A & Stream B & Implication \\ \midrule
Highway Cruise & Low & Low & Normal Monitoring \\
Uphill Towing & \textbf{Low} & \textbf{High} & Drivetrain Fatigue \\
Pothole/Shock & \textbf{High} & Low & Suspension/Chassis Risk \\
Rough Terrain & High & High & Immediate Inspection \\ \bottomrule
\end{tabular}
\end{table}

\section{Experimental Setup}
\label{sec:setup}

To validate the proposed architecture, we utilized a high-fidelity simulation environment to generate diverse operational contexts and an embedded RISC-V platform to verify runtime feasibility. Data were generated using the CARLA simulator \cite{Dosovitskiy:ACRL:2017} with heavy-duty vehicle dynamics (Firetruck model); 218 runs varied vehicle mass ($8300$--$13{,}500$,kg) and friction across contexts (normal roads, potholes, speed bumps, ramps). The dataset contains $N=192{,}857$ non-overlapping windows (30 samples, 10Hz), partitioned 70/15/15\% (train/val/test); splits were stratified by run ID to prevent data leakage, ensuring windows from the same mission do not cross splits, and mass and context distributions were balanced to prevent spurious correlations. Runtime profiling used a StarFive VisionFive 2 (JH7110, Quad-core U74 @ 1.5 GHz), measured as the incremental energy/latency over idle for 1{,}000 single-window (batch=1), single-threaded inferences using a KCX-017 meter.

We compare the proposed unsupervised architecture against a fully supervised benchmark: \textbf{Stream A (Unsupervised Baseline)} is the LSTM Autoencoder from Section \ref{sec:methods}, and the \textbf{Supervised Benchmark (XGBoost)} is a Gradient Boosting classifier mapping reconstruction-error features to discrete labels \cite{spotorno2025billing}. The supervised benchmark is constrained to the \textit{same feature space} as the unsupervised stream (reconstruction error statistics) to isolate feature performance; hyperparameters were optimized via grid search (maximizing weighted F1-score). Ground-truth labels (Normal, Pothole, Ramp, Rough Terrain) \cite{spotorno2025billing} reflect physical stress: low for cruising, high for transient shocks, and intermediate for sustained loads (e.g., ramps), capturing cumulative effort rather than just instability.

\section{Experimental Results and Analysis}
\label{sec:results}

This section first quantifies the limitations of the standalone unsupervised stream and then demonstrates the efficacy of the proposed Dual-Stream fusion.

\subsection{Validation of Macroscopic Physics Proxies}
We first validate the physics proxies as mechanical stress indicators independent of the ML stream, utilizing all $N=192{,}857$ windows. Regarding discriminative power, Mann--Whitney U tests ($p < 0.001$) confirm operational context separation: Suspension Stress distinguishes potholes ($\mu = 508.7$\,J) from normal driving ($\mu = 321.5$\,J); Drivetrain Stress separates ramp climbs ($\mu = 153.5$\,kJ) from level roads ($\mu = 136.7$\,kJ), consistent with sustained tractive effort. Additionally, Spearman correlation between vehicle mass ($8300$, $10{,}900$, $13{,}500$\,kg) and Drivetrain Stress ($\rho = 0.169$, $p < 0.001$) confirms monotonic capture of increased Newtonian work despite similar kinematics. These results establish the proxies as robust references for the subsequent correlation analysis.

\subsection{Analysis of Anomaly Detection Limitations}
\label{subsec:correlations}

\begin{figure}[ht!]
    \centering
    \includegraphics[width=\linewidth]{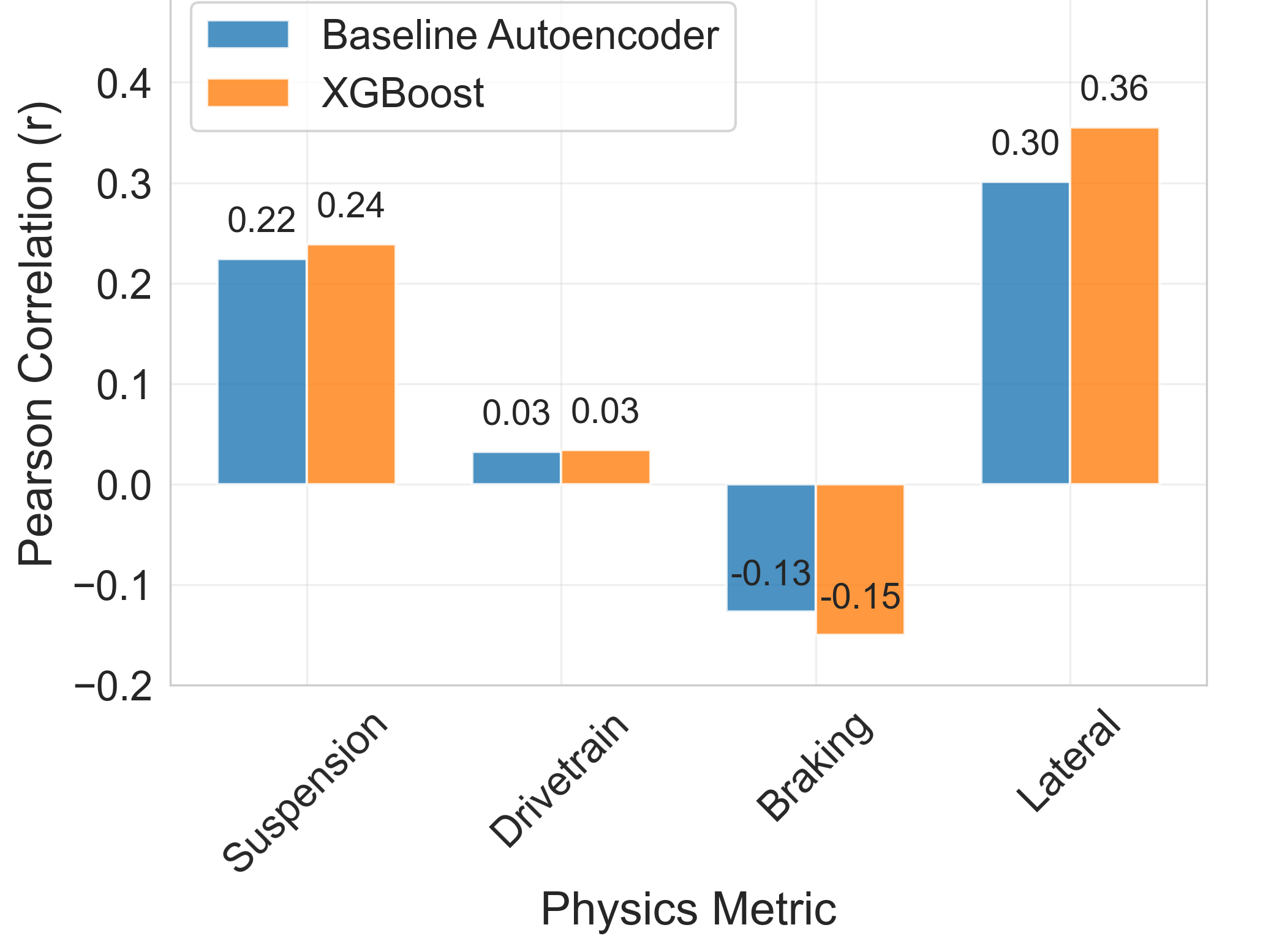}
    \caption{Correlation analysis ($N=192,857$ windows): Stream A correlates with lateral stability but decouples from drivetrain work ($r \approx 0$).}
    \label{fig:correlations}
\end{figure}

Stream A reliably separates events and preserves impact ordering (e.g., crash $ > $ pothole $ > $ ramps) \cite{spotorno2025billing}. However, correlating anomaly scores with physics proxies reveals intrinsic limitations. Fig.~\ref{fig:correlations} shows Stream A correlates with Lateral ($r=0.435$) and Suspension Stress ($r=0.218$) but decouples from Drivetrain ($r=0.032$) and Braking Energy ($r=-0.127$). This confirms the model functions as a \textit{surface stability monitor}, blind to Newtonian work. The effect worsens with payload: correlation with Drivetrain Stress drops from $r=0.085$ ($8300$\,kg) to $r=-0.007$ ($13{,}500$\,kg). Even the supervised XGBoost benchmark shows low correlation ($r=0.034$), indicating this limitation is fundamental to the signal characteristics of smooth, high-load events.

\begin{figure}[ht!]
    \centering
    \includegraphics[width=\linewidth]{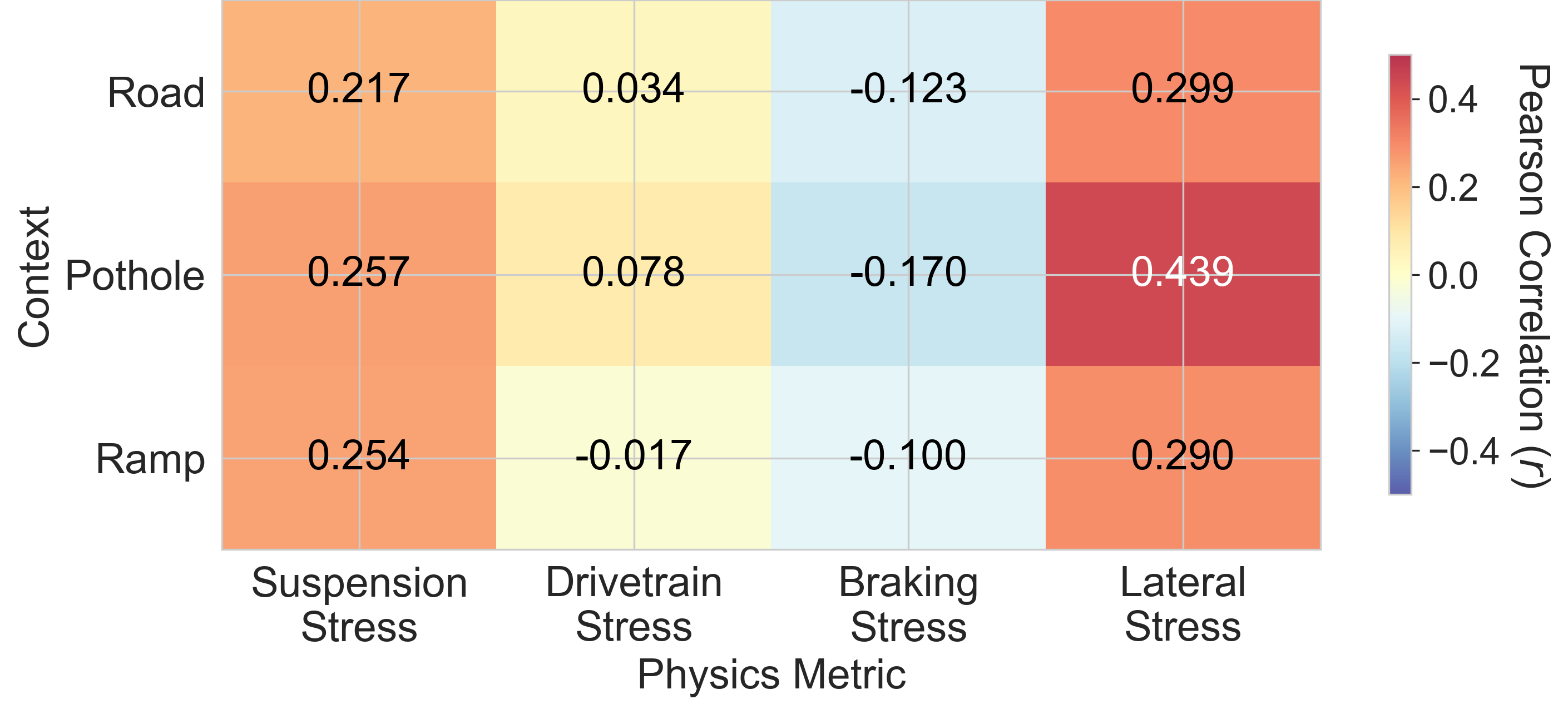}
    \caption{Scenario-specific correlations: negligible drivetrain correlation in Ramp events vs. strong instability capture in Potholes.}
    \label{fig:heatmap}
\end{figure}

Fig.~\ref{fig:heatmap} isolates this by scenario. Correlation with Lateral Stress is robust in \textit{Potholes} ($r=0.435$) but vanishes for Drivetrain Stress on \textit{Ramps} ($r=0.006$). Stream A fails to register high-intensity operations if they are signal-consistent (smooth). This results from the autoencoder's training on road data, including both smooth and agressive driving. Sustained forces (e.g., climbing) appear as consistent, low-error patterns, whereas transient shocks violate learned dynamics. The decoupling is not a model failure but a specialization toward surface instability. Besides, analyzing the behavior in high-stress regimes, by focusing on the top 10\% most energetic windows, Lateral Stress correlation remains strong ($r = 0.368$), while Suspension and Drivetrain correlations degrade to noise ($r \approx -0.1$). Repetitive high-intensity signals are absorbed into the learned notion of normality, reinforcing the model's focus on transient instability over cumulative fatigue.

\subsection{Evaluation of the Dual-Stream Architecture}
\begin{figure}[htbp]
    \centering
    \includegraphics[width=\linewidth]{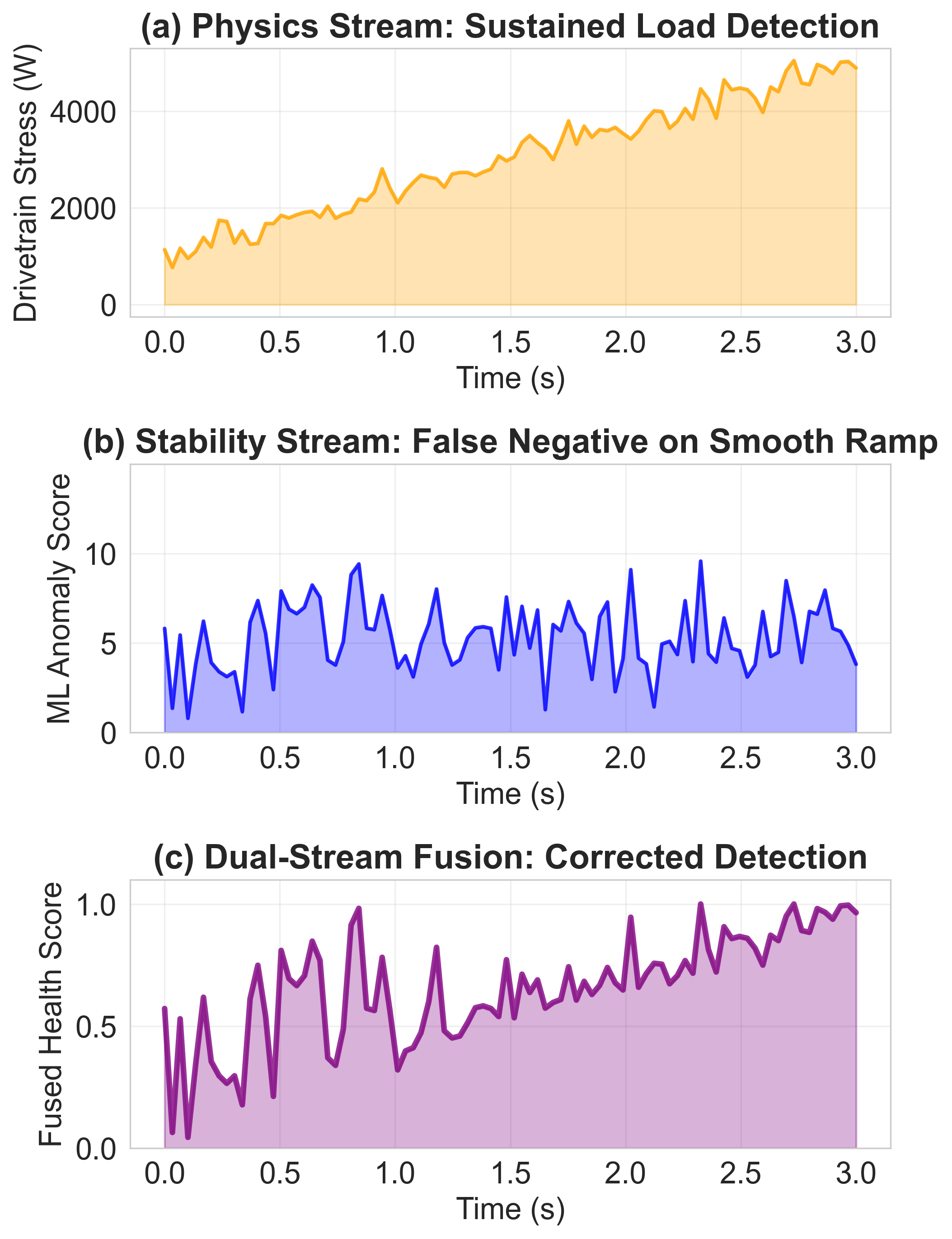}
    \caption{Dual-stream fusion in ramp climb: (a) physics stream detects load, (b) ML stream misses it, (c) fused output correctly flags high intensity.}
    \label{fig:fusion_solution}
\end{figure}

The proposed fusion logic addresses this blind spot through architectural orthogonality. As shown in Section \ref{subsec:correlations}, the correlation between the ML reconstruction error and the Physics proxies is near zero. This statistical independence implies that the two streams capture non-overlapping regions of the vehicle's operational envelope: Stream A captures high-frequency instability, while Stream B captures low-frequency work. Therefore, the Max-Pooling fusion is not merely a heuristic but a logical union of the state space. Fig \ref{fig:fusion_solution} illustrates this additive coverage: the "Ramp Climb" generates negligible ML error (False Negative) but high Drivetrain Stress. A single fused metric is not strictly necessary, operators may monitor streams individually, but the fused score provides a single "care-needed" index that guarantees the superset of wear-inducing events is captured.

\subsection{Computational Feasibility Analysis}
Table \ref{tab:compute} details the computational overhead of adding Stream B on the RISC-V platform. Since the proxies require only basic arithmetic, they add negligible overhead ($\approx$1.3\% time/energy; within measurement variance). The Dual-Stream architecture resolves the blind spot with minimal penalty, validating its suitability for runtime monitoring on low-power ECUs.

\begin{table}[t]
\caption{Computational Cost on RISC-V (StarFive JH7110)\protect\footnotemark}
\label{tab:compute}
\centering
\begin{tabular}{@{}lccc@{}}
\toprule
\textbf{Component} & \textbf{Time ($\pm\sigma$) ($\mu$s)} & \textbf{Energy ($\pm\sigma$) ($\mu$J)} \\
\midrule
Stream A (ML Inference) & 6.268 (0.132) & 0.671 (0.047) \\
Stream B (Physics Proxies) & $\approx$ 0.08 (0.04) & $\approx$ 0.009 (0.005) \\
\midrule
\textbf{Total Dual-Stream} & \textbf{6.35 (0.14)} & \textbf{0.680 (0.049)} \\
\bottomrule
\end{tabular}
\end{table}
\footnotetext{Energy and time values represent incremental consumption above the idle baseline. Stream B values are computed as the difference between full dual-stream and Stream A measurements; small variances reflect profiling noise.}

\section{Discussion and Conclusion}
\label{sec:discussion_conclusion}

This paper demonstrates that unsupervised anomaly detection alone is insufficient for comprehensive vehicle operational effort estimation, highlighting a fundamental limitation in prevailing data-driven monitoring: statistical surprise does not equate to mechanical burden. By validating against macroscopic physics on large-scale sensor streams, we quantify this blind-spot, showing how a reconstruction-based Autoencoder baseline robustly detects lateral instability ($r=0.435$) but ignores sustained drivetrain work ($r \approx 0$).

To resolve these load-blind scenarios, the proposed Dual-Stream Architecture integrates unsupervised learning for dynamic hazards with macroscopic physics proxies for cumulative load. By using proxies as a parallel validator rather than training constraints, this hybrid approach preserves unsupervised label efficiency. From a practical engineering perspective, this avoids the prohibitive data annotation and computational costs of purely supervised models. Crucially, it leverages existing sensors rather than full physics models requiring unavailable telemetry \cite{Wong2022}. Computational analysis demonstrates feasibility on embedded RISC-V platforms \cite{spotorno2025billing} with an energy consumption of approximately 0.68 $\mu$J per full dual-stream inference (LSTM Autoencoder plus physics proxies), enabling trustworthy, on-device condition monitoring on standard, resource-constrained ECUs.

We explicitly address the reality of sensor noise in real-world deployments. The macroscopic proxies rely on numerical differentiation of low-frequency (10\,Hz) signals, which can amplify noise. We mitigate this by applying a symmetric central difference followed by a short 5-point moving average, retaining zero-phase derivative estimation while reducing variance. While this approach does not recover high-frequency micro-vibration content (aliasing risk for $>$5\,Hz components), the proxies are intended and validated specifically as macroscopic exposure measures (sustained tractive work, braking dissipation, and low-frequency suspension events) rather than strain-level fatigue surrogates.

Regarding the drivetrain proxy, sensitivity to parameters such as mass $m(t)$ is a deterministic feature of physics-based monitoring, not a robustness flaw. Unlike black-box models with unpredictable sensitivities, our proxy's linear dependence on mass ensures a heavier vehicle \textit{correctly} registers proportionally higher stress. While absolute precision requires accurate parameter estimation or conservative upper-bound estimates, this guarantees that relative trends critical for predictive maintenance remain physically consistent. As reported in Section~\ref{sec:results}, the architecture monotonically captures increasing work with mass.

By quantifying actual mechanical burden rather than relying on static mileage, this approach provides the granular data required to shift toward dynamic condition-based maintenance. This addresses critical pain points in municipal and commercial fleet management, where traditional scheduling fails to capture the high variance in operational intensity \cite{mattetti2022canbus}, because fleets consist of diverse operational profiles \cite{sayedchallenges}, thus uniform maintenance policies remain highly inefficient \cite{kazemian2024sensor}. Furthermore, compared to baseline alternatives like clustered anomaly scoring or GIS-based indices, our dual-stream approach provides continuous, on-board multi-dimensionality without external data dependencies. By maintaining separate streams for instability and load, we preserve the interpretability necessary for stakeholder trust, directly addressing growing regulatory concerns regarding black-box AI in industrial systems \cite{perez2024artificial}.

Limitations of this study include reliance on simulated data, low-frequency sampling bounds on micro-event precision, and the necessity of parameter assumptions. Future work must focus on real-vehicle deployment for broader validation, quantifying the propagation of mass and drag uncertainty to absolute wear estimates, and developing adaptive proxy calibration, such as estimating mass directly from kinematics, to ensure this explainable health vector adapts to evolving operational contexts.

\bibliographystyle{IEEEtran}
\bibliography{references}

\end{document}